\documentclass{article}
\usepackage{spconf,amsmath,graphicx}
\usepackage{subfigure}
\usepackage{enumitem}

\usepackage[pagebackref=true,breaklinks=true,letterpaper=true,colorlinks,bookmarks=false]{hyperref}

\usepackage{gensymb,graphics,subfigure,inputenc}
\usepackage[linesnumbered,algo2e,boxed]{algorithm2e}
\graphicspath{{Figure/}}
\usepackage[figuresright]{rotating}
\usepackage{amsmath,amssymb,textcomp,subfigure,multirow,upgreek}
\usepackage{booktabs}
\usepackage{color,mdwlist}

\usepackage{caption}
\title{Exocentric to Egocentric Image Generation via Parallel Generative Adversarial Network}
\name{Gaowen Liu$^1$, Hao Tang$^{1,2}$, Hugo Latapie$^{3}$, Yan Yan$^1$\sthanks{Corresponding author.}}
\address{$^1$Department of Computer Science, Texas State University, USA\\
	$^2$DISI, University of Trento, Italy\\
	$^3$Chief Technology \& Architecture Office, Cisco, USA
	}

\begin{document}
%
\maketitle
\begin{abstract}
Cross-view image generation has been recently proposed to generate images of one view from another dramatically different view. In this paper, we investigate exocentric (third-person) view to egocentric (first-person) view image generation. This is a challenging task since egocentric view sometimes is remarkably different from exocentric view. Thus, transforming the appearances across the two views is a non-trivial task. To this end, we propose a novel Parallel Generative Adversarial Network (P-GAN) with a novel cross-cycle loss to learn the shared information for generating egocentric images from exocentric view. We also incorporate a novel contextual feature loss in the learning procedure to capture the contextual information in images. Extensive experiments on the Exo-Ego datasets~\cite{third2019} show that our model outperforms the state-of-the-art approaches. 
\end{abstract}
\begin{keywords}
Egocentric, Exocentric, Cross-View Image Generation, Parallel GANs
\end{keywords}
\section{INTRODUCTION}
\label{sec:intro}
\vspace{-0.3cm}
Wearable cameras, also known as first-person cameras, nowadays are widely used in our daily lives since the appearance of low price but high quality wearable products such as GoPro cameras. Meanwhile, egocentric (first-person) vision is also becoming a critical research topic in the field. As we know, egocentric view have some unique properties other than exocentric (third-person) view. Traditional exocentric cameras usually give a wide and global view of the high-level appearances happened in a video. However, egocentric cameras can capture the objects and people at a much finer level of granularity. In the early egocentric vision studies, researchers \cite{kanade} found that people perform different activities or interacting with objects from a first-person egocentric perspective and seamlessly transfer knowledge between egocentric and exocentric perspective. Therefore, analyzing the relationship between egocentric and exocentric perspectives is an extremely useful and interesting topic for image and video understanding. However, there is few research to address this important problem in literature.


\begin{figure*}[htbp]
\center
{\includegraphics[width=0.8\linewidth]{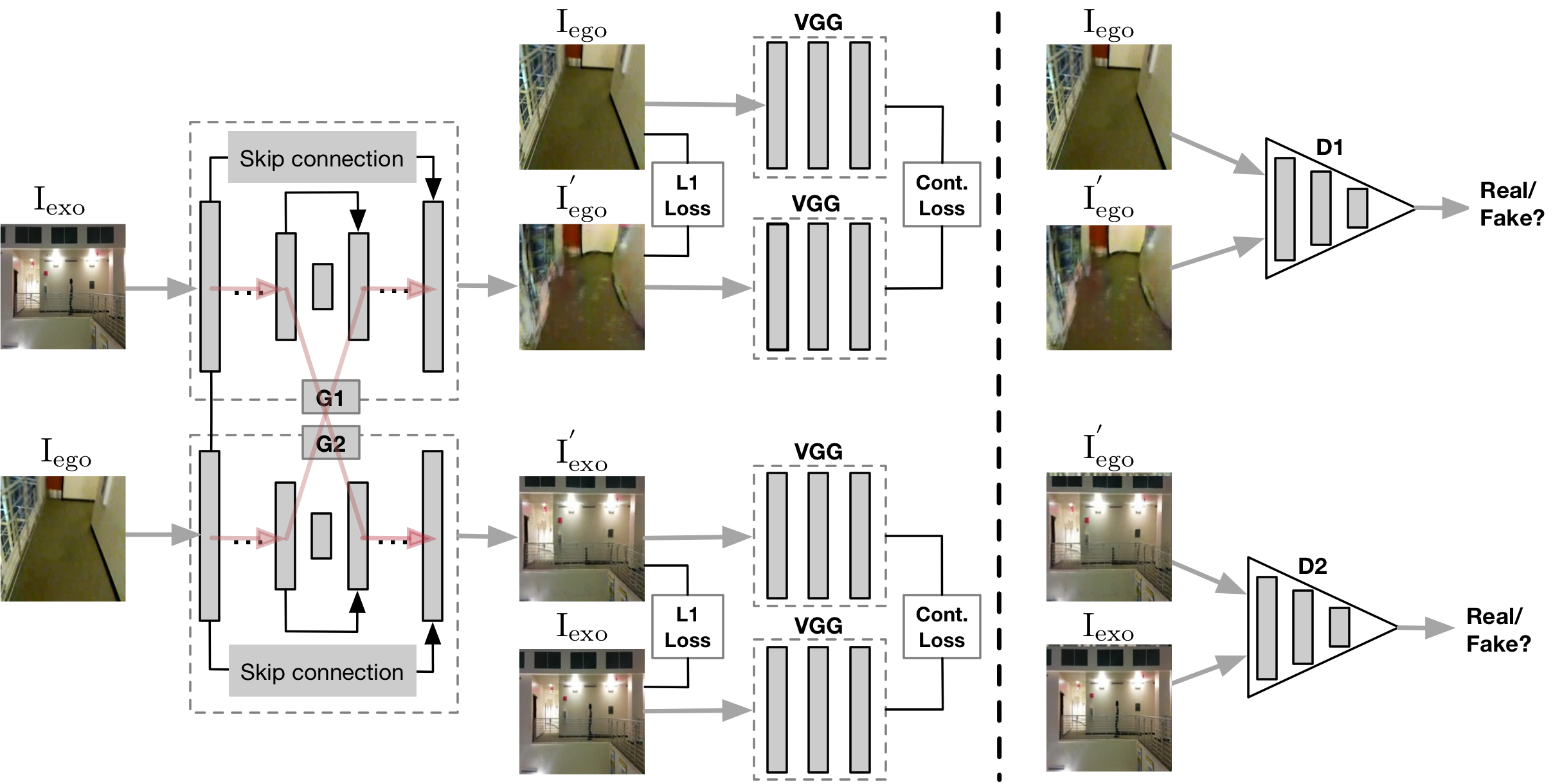}} 
\caption{The pipeline of our P-GAN model. It consists of two parallel generators ($G_1$, $G_2$) and two discriminators ($D_1$, $D_2$). The total loss contains pairs of $L1$ loss, contextual loss and adversarial loss.}
\vspace{-0.4cm}
\label{fig:architecture}
\end{figure*}

Recently, Generative Adversarial Networks (GANs) \cite{goodfellow2014generative} have been shown effectively in image generation tasks. Isola \textit{et al.}~\cite{isola2017image} propose Pix2Pix adversarial learning framework on paired image generation, which is a supervised model and uses a conditional GAN framework to learn a translation function from input to output image domain. Zhu \textit{et al.}~\cite{CycleGAN2017} introduce CycleGAN which develops cycle-consistency constraint to deal with unpaired image generation. However, these existing works consider an application scenario in which the objects and the scenes have a large degree of overlapping in appearance and view. Recently, some works investigate cross-view image generation problems to generate a novel scene which is drastically different from a given scene image. This is a more challenging task since different views share little overlap information. To tackle this problem, Regmi and Borji~\cite{Regmi_2018_CVPR} propose X-Fork and X-Seq GAN-based architecture using an extra semantic segmentation map to facilitate the generation. Moreover, Tang \textit{et al.}~\cite{tang2019multichannel} propose a multi-channel attention selection module within a GAN framework for cross-view image generation. However, these methods are not able to generate satisfactory results due to the drastically differences between exocentric and egocentric views.

To bridge egocentric and exocentric analaysis, in this paper we propose a novel Parallel GAN (P-GAN) to generate exocentric images from egocentric view. P-GAN framework is able to automatically learn the shared information between two parallel generation tasks via a novel cross-cycle loss and hard-sharing of network layers. We also utilize a novel contextual loss in our objective function to capture texture information over the entire images. To the best of our knowledge, we are the first to attempt to incorporate a parallel generative network for exocentric to egocentric image translation. 
The proposed P-GAN is related to CoGAN~\cite{DBLP:journals/corr/0001T16} and DualGAN~\cite{DualGAN}. However, CoGAN and DualGAN have limited ability in generating image pairs with dramatically different viewpoints. As shown in Fig.~\ref{fig:architecture}, our architecture is designed in a bi-directional parallel fashion to discover the shared information between egocentric and exocentric images. Two parallel GANs are trained simultaneously with hard-sharing of certain layers.




In summary, our contributions can be highlighted as follows. (i) A novel P-GAN is proposed to learn the shared information between different views simultaneously via a novel cross-cycle loss. (ii) A novel contextual feature loss is incorporated in the training to capture the contextual information. (iii) Experiments on Exo-Ego dataset show the effectiveness of our hard-sharing of network layers in multi-directional parallel generative models. 


\section{Parallel GANs}
\vspace{-0.3cm}
\label{sec:format}

\subsection{Network Architecture}
\vspace{-0.3cm}
\label{ssec:subhead}
Cross-view exocentric to egocentric image synthesis is a challenging task, because these two views have little overlapping in image appearance. Most existing works on cross-view image synthesis are based on GANs. A traditional GAN consists of a generative model and a discriminative model. The objective of the generative model is to synthesize images resembling real images, while the objective of the discriminative model is to distinguish real images from synthesized ones. Both the generative and discriminative models are realized as multi-layer perceptrons. Since there will be some shared high-level concept information in a pair of corresponding images between exocentric and egocentric views, we propose a P-GAN with two GANs in parallel which is able to learn the shared high-level semantic information among different views. Fig.~\ref{fig:architecture} shows our framework which contains two generators and two discriminators. A set number of layers from two generators are shared across P-GAN. We force the first three layers of two generators to have the identical structure and share the weights, and the rest layers are task-specific. The experiments show that sharing three layers of generators yield the best performance.

Particularly, we employ U-Net \cite{Unet} as the architecture of our generators $G_1$ and $G_2$. We impose skip connection strategy from down-sampling path to up-sampling path to avoid vanishing gradient problem. To learn the shared information between exocentric and egocentric view, we perform hard-sharing in the first three layers of down-sampling path. We adopt PatchGAN \cite{isola2017image} for the discriminator $D_1$ and $D_2$. The feature maps for contextual loss are extracted by the VGG-19 network pretrained on ImageNet.

\begin{figure*}[htbp] \small
\center
{\includegraphics[width=0.8\linewidth]{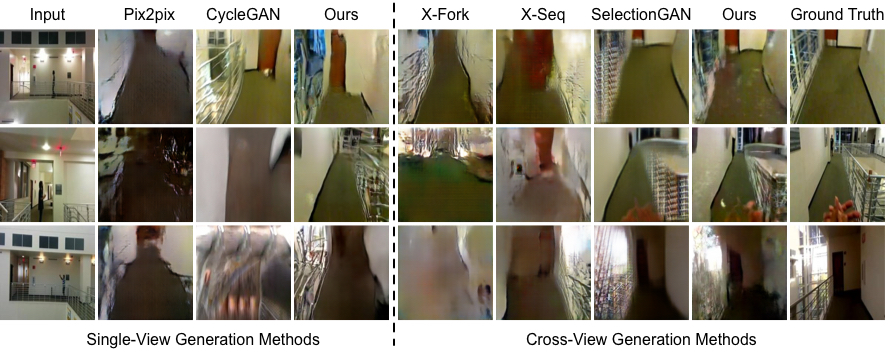}} 
\caption{Results generated by different methods on Side2Ego dataset. These samples were randomly selected for visualization purposes. Columns from left to right are: Input, Pix2pix \cite{isola2017image}, CycleGAN \cite{CycleGAN2017}, P-GAN (ours), X-Fork \cite{Regmi_2018_CVPR}, X-Seq \cite{Regmi_2018_CVPR}, SelectionGAN \cite{tang2019multichannel}, P-GAN + Segmentation map (ours), Ground Truth. }
\vspace{-0.4cm}
\label{fig:fig1}
\end{figure*}
\vspace{-0.2cm}
\subsection{Overall Optimization Objective}
\label{ssec:subhead}
\vspace{-0.2cm}
The training objective can be decomposed into four main components which are contextual loss, adversarial loss, cross-cycle loss and reconstruction loss.

\noindent \textbf{Contextual loss.} Different from the commonly used $L1$ loss function which compares pixels at the same spatial coordinates between the generated image and the target image, we incorporate contextual loss in our P-GAN learning framework. The key idea is to measure similarity between images at the high-level feature space. 

Given a generated fake image $I'_{ego}$ and a real image $I_{ego}$ in egocentric view, we obtain a list of VGG-19 \cite{vgg} features as $I_{ego} {=} \{I_i\}$ and $I'_{ego} {=} \{I'_j\}$, where $I_i {=} \psi^i(I_{ego})$, $I'_j {=} \psi^j(I'_{ego})$, $\psi$ means VGG-19 feature. $i$, $j$ are $i$-th and $j$-th layer in the network $\psi$. The similarity between the generated image $I'_{ego}$ and the real image $I_{ego}$ in egocentric view can be defined as follows,

\vspace{-0.1cm}
\begin{equation}
\begin{aligned}
\mathcal{S}_{I_i,I'_j} = \mathrm{exp} \left( 1-\frac{1-d_{ij}}{\mathrm{min}_k d_{ik} + \zeta} \right)/h
\end{aligned}
\label{eqn:Similarity}
\vspace{-0.1cm}
\end{equation}
where $d_{ij}$ is the cosine distance between $I_{ego}$ and $I'_{ego}$. We define $\zeta {=} 1e^{-5}$, $h {=} 0.5$ in our experiments. The similarity can be normalized as, 
\vspace{-0.1cm}
\begin{equation}
\begin{aligned}
\bar{S}_{ij} = \frac{\mathcal{S}_{I_i,I'_j}}{\sum_{k}\mathcal{S}_{I_i,I'_k}}
\end{aligned}
\label{eqn:contij}
\vspace{-0.1cm}
\end{equation}
Then the contextual loss is formulated as follows,
\vspace{-0.1cm}
\begin{equation}
\begin{aligned}
\mathcal{L}_{cont}(I_i,I'_j) = \frac{1}{\mathrm{max}(\mid I_{ego} \mid, \mid I'_{ego} \mid )}\sum_j \mathrm{max}~\bar{S}_{ij}
\end{aligned}
\label{eqn:contextual}
\vspace{-0.1cm}
\end{equation}
where $|\cdot|$ denotes the numbers of feature maps.

\noindent \textbf{Cross-cycle loss.} As shown in Fig.~\ref{fig:architecture}, we employ U-Net \cite{Unet} as our generators $G_1$ and $G_2$. Each U-Net contains a down-sampling encoder $EN$ which is a feature contracting path, and an up-sampling decoder $DE$ which is a feature expanding path. Inspired by the U-net properties, we design a novel cross-cycle loss as follows,
\vspace{-0.1cm}
\begin{equation}
\begin{aligned}
\mathcal{L}_{X}(G_1, G_2) =   
& \mathbb{E}_{I_{exo}, I^{'}_{exo}} \left[ \|I_{exo}-DE_2(EN_1(I_{exo}))\|_1 \right] + \\ 
& \lambda_1\mathbb{E}_{I_{ego}, I^{'}_{ego}} \left[ \|I_{ego}-DE_1(EN_2(I_{ego}))\|_1 \right]
\end{aligned}
\label{eqn:LX}
\vspace{-0.1cm}
\end{equation}

\noindent \textbf{Adversarial loss.} Recent works \cite{goodfellow2014generative,salimans2016improved,liu2018improved,che2016mode,tang2019cycle} have shown that one can learn a mapping function by tuning a generator and a discriminator in an adversarial way. Assuming we target to learn a mapping $G\colon I_{exo} {\to} I_{ego}$ from input exocentric image $I_{exo}$ to output egocentric image $I_{ego}$. The generator $G$ is trained to produce outputs to fool the discriminator $D$. 
The adversarial loss can be expressed as,
\vspace{-0.1cm}
\begin{equation}
\begin{aligned}
\mathcal{L}_{GAN_1}(G_1,D_1) =  
& \mathbb{E}_{I_{exo}, I_{ego}} \left[ \log D_1(I_{exo}, I_{ego}) \right] +  \\
& \mathbb{E}_{I_{exo}, I^{'}_{ego}} \left[\log (1 - D_1(I_{exo}, G_1(I_{exo}))) \right]
\end{aligned}
\label{eqn:adv_1}
\vspace{-0.1cm}
\end{equation}
\vspace{-0.1cm}
\begin{equation}
\begin{aligned}
\mathcal{L}_{GAN_2}(G_2,D_2) =  
& \mathbb{E}_{I_{ego}, I_{exo}} \left[ \log D_2(I_{ego}, I_{exo}) \right] +  \\
& \mathbb{E}_{I_{ego}, I^{'}_{exo}} \left[\log (1 - D_2(I_{ego}, G_2(I_{ego}))) \right]
\end{aligned}
\label{eqn:adv_2}
\vspace{-0.1cm}
\end{equation} 
The adversarial loss is the sum of Eq.~\eqref{eqn:adv_1} and Eq.~\eqref{eqn:adv_2}.
\vspace{-0.1cm}
\begin{equation}
\begin{aligned}
\mathcal{L}_{GAN} = \mathcal{L}_{GAN_1}(G_1,D_1) + \lambda_2\mathcal{L}_{GAN_2}(G_2,D_2)
\end{aligned}  
\label{eqn:adv}
\vspace{-0.1cm}
\end{equation}

  

\noindent \textbf{Reconstruction loss.} The task of the generator is to reconstruct an image as close as the target image. We use $L1$ distance in the reconstruction loss,
\vspace{-0.1cm}
\begin{equation}
\begin{aligned}
\mathcal{L}_{re}(G_1, G_2) =  
& \mathbb{E}_{I_{exo}, I^{'}_{ego}} \left[ \|I_{ego}-DE_1(EN_1(I_{exo}))\|_1 \right] + \\
& \lambda_3\mathbb{E}_{I_{ego}, I^{'}_{exo}} \left[ \|I_{exo}-DE_2(EN_2(I_{ego}))\|_1 \right]
\end{aligned}
\label{eqn:L1}
\vspace{-0.1cm}
\end{equation}

%

\begin{figure*}[t] \small
\center
{\includegraphics[width=0.8\linewidth]{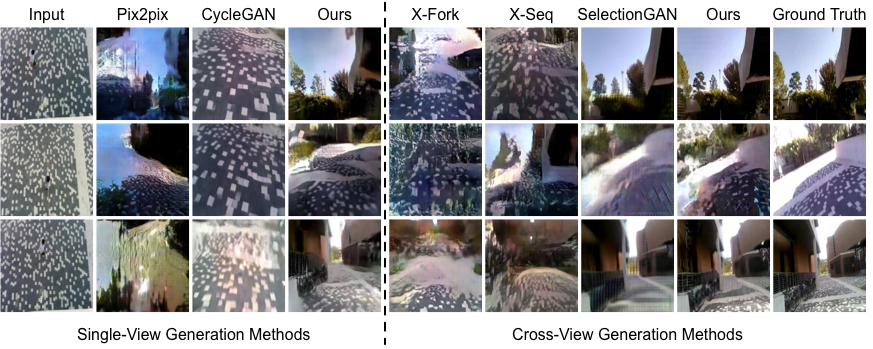}}
\caption{Results generated by different methods on Top2Ego dataset. These samples were randomly selected for visualization purposes. Columns from left to right are: Input, Pix2pix \cite{isola2017image}, CycleGAN \cite{CycleGAN2017}, P-GAN (ours), X-Fork \cite{Regmi_2018_CVPR}, X-Seq \cite{Regmi_2018_CVPR}, SelectionGAN \cite{tang2019multichannel}, P-GAN + Segmentation map (ours), Ground Truth.}
\label{fig:fig2}
\vspace{-0.4cm}
\end{figure*}

\noindent \textbf{Overall loss.} The total optimization loss is a weighted sum of the above losses. Generators $G_1$, $G_2$ and discriminators $D_1$, $D_2$ are trained in an end-to-end fashion to optimize the following objective function,
\vspace{-0.1cm}
\begin{equation}
\begin{aligned}
\mathcal{L} =  \mathcal{L}_{GAN} + \lambda_4\mathcal{L}_X + \lambda_5\mathcal{L}_{re} + \lambda_6\mathcal{L}_{cont}
\end{aligned}
\label{eqn:overall}
\end{equation}
where $\lambda _i$'s are the regularization parameters.

\begin{table*}[!t] \small
 \centering
 \caption{SSIM, PSNR, Sharpness Difference (SD), KL score (KL) and Accuracy of different single-view image generation methods. For these metrics except KL score, higher is better.}
 \resizebox{0.62\linewidth}{!}{%
  \begin{tabular}{cccccccccc} \toprule
   \multirow{2}{*}{Dataset} & \multirow{2}{*}{Method}  &  \multirow{2}{*}{SSIM} & \multirow{2}{*}{PSNR} & \multirow{2}{*}{SD} & \multirow{2}{*}{KL} & \multicolumn{2}{c}{Top-1} & \multicolumn{2}{c}{Top-5} 
   \\&&&&&& \multicolumn{2}{c}{Accuracy (\%)} & \multicolumn{2}{c}{Accuracy (\%)}  \\ \hline
   \multirow{3}{*}{Top2Ego} & Pix2pix\cite{isola2017image} &0.2514 & 15.0532 & 18.1002 & 62.74 $\pm$ 1.78 & 1.24 & 1.22 & 4.21 & 4.35 \\
   & CycleGAN \cite{CycleGAN2017}   & 0.2806 &15.5486 & 18.5678 & 52.09 $\pm$ 1.69 & \textbf{2.10} & 0.99 & 5.37 & 2.72\\ 
   & Ours  & \textbf{0.3098} & \textbf{17.0236} & \textbf{18.6043} & \textbf{31.46} $\pm$ \textbf{1.74} & 1.81 & \textbf{5.90} & \textbf{5.74} & \textbf{9.17} \\\hline
   \multirow{3}{*}{Side2Ego} & Pix2pix \cite{isola2017image}         &0.3946 & 16.0716 & 19.8664 & 75.27 $\pm$ 2.01 & 3.20 & 5.18 & 8.41 & 13.30 \\
   & CycleGAN \cite{CycleGAN2017}  & 0.4017 & 15.9678 & 19.7533 & 62.41 $\pm$ 2.41 & 4.18 & 7.60 & 15.62 & 21.45\\
   & Ours & \textbf{0.4908} & \textbf{17.995}1 & \textbf{20.6521} & \textbf{13.92 $\pm$ 1.53} & \textbf{16.21} & \textbf{30.80} &\textbf{27.57} &\textbf{46.51}\\
   \bottomrule  
 \end{tabular}}
 \label{tab:single}
 \vspace{-0.4cm}
\end{table*}
\begin{table*}[!t] \small
 \centering
 \caption{SSIM, PSNR, Sharpness Difference (SD), KL score (KL) and Accuracy of different cross-view image generation methods. For these metrics except KL score, higher is better.}
 \resizebox{0.62\linewidth}{!}{%
  \begin{tabular}{cccccccccc} \toprule
   \multirow{2}{*}{Dataset} & \multirow{2}{*}{Method}  &  \multirow{2}{*}{SSIM} & \multirow{2}{*}{PSNR} & \multirow{2}{*}{SD} & \multirow{2}{*}{KL} & \multicolumn{2}{c}{Top-1} & \multicolumn{2}{c}{Top-5} 
   \\&&&&&& \multicolumn{2}{c}{Accuracy (\%)} & \multicolumn{2}{c}{Accuracy (\%)}  \\ \hline
   \multirow{4}{*}{Top2Ego} & X-Fork \cite {Regmi_2018_CVPR} &0.2952 & 15.8849 & 18.7349 & 63.96$\pm$1.74&0.8 & 1.22 & 3.16 &4.08 \\
   & X-Seq \cite {Regmi_2018_CVPR}  &0.3522 & 16.9439 & 19.2733 & 54.91 $\pm$ 1.81 & 1.07 & 1.77 & 4.29 & 6.94\\ 
   & SelectionGAN \cite{tang2019multichannel}  & 0.5047 & 22.0244 & 19.1976 & \textbf{10.07 $\pm$ 1.29} & 8.85 & 16.55 & 24.32 & 33.90\\
   & Ours  & \textbf{0.5287} & \textbf{22.2891} & \textbf{19.2389} & 12.07 $\pm$ 1.69 & \textbf{9.76} & \textbf{29.67} & \textbf{24.80} & \textbf{51.79}
   \\\hline
   \multirow{4}{*}{Side2Ego} & X-Fork \cite{Regmi_2018_CVPR}         &0.4499 & 17.0743 & 20.4443 & 51.20 $\pm$ 1.94 & 4.49 & 9.76 & 11.63& 19.44 \\
   & X-Seq \cite{Regmi_2018_CVPR}  & 0.4763 & 17.1462 & 20.7468 & 45.10 $\pm$ 1.95 & 6.51 & 12.70 & 11.97 & 19.36\\
   & SelectionGAN \cite{tang2019multichannel}  & 0.5128 & 18.3021 & 20.9426 & \textbf{7.26 $\pm$ 1.27} & 20.84 & 37.49 & 42.51 & 65.22\\
   & Ours & \textbf{0.5205} & \textbf{19.4521} & \textbf{20.9684} & 25.25 $\pm$ 1.88 & \textbf{20.96} & \textbf{39.08} &\textbf{42.58} &\textbf{66.00}\\
   \bottomrule  
 \end{tabular}}
 \label{tab:cross}
 \vspace{-0.4cm}
\end{table*}

\section{Experimental Results}
\vspace{-0.3cm}
\label{sec:pagestyle}
\textbf{Datasets.} To explore the effectiveness of our proposed P-GAN model, we compare our model with the state-of-the-art methods on Exo-Ego dataset \cite{third2019} which contains two different viewpoint subsets (Side2Ego and Top2Ego). This dataset is challenging due to two reasons. First, it contains dramatically different indoor and outdoor scenes. Second, the dataset is collected simultaneously by an exocentric camera (side and top view) and an egocentric body-worn wearable camera. It includes a huge amount of blurred images for egocentric view. For Side2Ego subset, there are 26,764 pairs of images for training and 13,788 pairs for testing. For Top2Ego subset, there are 28,408 pairs for training and 14,064 pairs for testing. All images are in high-resolution $1280 {\times} 720$ pixels. 

\noindent \textbf{Experimental Setup.}
We compare our P-GAN with both single-view image generation methods~\cite{isola2017image, CycleGAN2017} and cross-view image generation methods~\cite{Regmi_2018_CVPR,tang2019multichannel}. We adopt the same experimental setup as in~\cite{isola2017image,Regmi_2018_CVPR,tang2019multichannel}. All images are scaled to $256 {\times} 256$. We enable image flipping and random crops for data augmentation. To compute contextual loss, we follow \cite{mechrez2018contextual} and use the VGG-19 network to extract image feature maps pretrained on ImageNet. 
We train 35 epochs with the batch size of 4. In our experiments, we set $\lambda_1 {=} 10$, $\lambda_2 {=} 10$, $\lambda_3 {=} 100$, $\lambda_4 {=} 10$, $\lambda_5 {=} 1$, $\lambda_6 {=} 1$ in Eq.~\eqref{eqn:LX}, \eqref{eqn:adv}, \eqref{eqn:L1} and \eqref{eqn:overall}, respectively. 
The state-of-the-art cross-view generation methods, \textit{i.e.}, X-Fork \cite{Regmi_2018_CVPR}, X-Seq \cite{Regmi_2018_CVPR} and SelectionGAN \cite{tang2019multichannel} utilize segmentation map to facilitate target view image generation. To compare with these cross-view methods, we adopt RefineNet~\cite{lin2019refinenet,Lin:2017:RefineNet} to generate segmentation maps on Side2Ego and Top2Ego subsets as in~\cite{Regmi_2018_CVPR,tang2019multichannel}. The generated segmentation maps are used as the conditional input of $G_1$ and $G_2$. 
The proposed P-GAN is implemented using PyTorch.

\noindent \textbf{Evaluation Metrics.}
We apply metrics such as top-k prediction accuracy and KL score for evaluations as in~\cite{tang2019multichannel,Regmi_2018_CVPR}. We also employ pixel-level similarity metrics, \textit{i.e.}, Structural-Similarity (SSIM), Peak Signal-to-Noise Ratio (PSNR) and Sharpness Difference (SD). These metrics evaluate the generated images in a high-level feature space. 

\noindent \textbf{Quantitative Results.} The quantitative results are presented in Tables~\ref{tab:single} and~\ref{tab:cross}. We observe that our P-GAN achieves better results than state-of-the-art methods in most cases. Compared with single-view image generation methods, our P-GAN outperforms Pix2pix \cite{isola2017image} and CycleGAN \cite{CycleGAN2017}. On the other hand, we also achieve better results than other cross-view image generation methods in most metrics while incorporating semantic segmentation map as in the SelectionGAN \cite{tang2019multichannel}.

\noindent \textbf{Qualitative Results.} Qualitative results are shown in Fig.~\ref{fig:fig1} and Fig.~\ref{fig:fig2}. The results confirm that the proposed P-GAN network has the ability to transfer the image representations from exocentric to egocentric view, \textit{i.e.,}  objects are in the correct positions for generated egocentric images. Results show that egocentric images generated by P-GAN are visually much better compared with other baselines.

\vspace{-0.2cm}
\section{CONCLUSIONS}
\vspace{-0.2cm}
\label{sec:typestyle}
In this paper, we introduce a novel P-GAN which is able to learn shared information between cross-view images via a novel cross-cycle loss for a challenging exocentric to egocentric view image generation task. 
Moreover, we incorporate a novel contextual feature loss to capture the contextual information in images. 
Experimental results demonstrate that the hard-sharing of network layers in multi-directional parallel generative models can be used to increase the performance of cross-view image generation.

\noindent \textbf{ACKNOWLEDGEMENT.}
This research was partially supported by a gift donation from Cisco Inc. and NSF NeTS-1909185 and NSF CSR-1908658. This article solely reflects the opinions and conclusions of its authors and not the funding agents.

\vfill\pagebreak





\bibliographystyle{IEEEbib}
\bibliography{Template}

\begin{thebibliography}{10}

\bibitem{third2019}
Mohamed Elfeki, Krishna Regmi, Shervin Ardeshir, and Ali Borji,
\newblock ``From third person to first person: Dataset and baselines for
  synthesis and retrieval,''
\newblock in {\em CVPR}, 2019.

\bibitem{kanade}
Takeo Kanade and Martial Hebert,
\newblock ``First-person vision,''
\newblock {\em Proceedings of the IEEE}, vol. 100, no. 8, pp. 2442--2453, 2012.

\bibitem{goodfellow2014generative}
Ian Goodfellow, Jean Pouget-Abadie, Mehdi Mirza, Bing Xu, David Warde-Farley,
  Sherjil Ozair, Aaron Courville, and Yoshua Bengio,
\newblock ``Generative adversarial nets,''
\newblock in {\em NeurIPS}, 2014.

\bibitem{isola2017image}
Phillip Isola, Jun-Yan Zhu, Tinghui Zhou, and Alexei~A Efros,
\newblock ``Image-to-image translation with conditional adversarial networks,''
\newblock in {\em CVPR}, 2017.

\bibitem{CycleGAN2017}
Jun-Yan Zhu, Taesung Park, Phillip Isola, and Alexei~A Efros,
\newblock ``Unpaired image-to-image translation using cycle-consistent
  adversarial networkss,''
\newblock in {\em ICCV}, 2017.

\bibitem{Regmi_2018_CVPR}
Krishna Regmi and Ali Borji,
\newblock ``Cross-view image synthesis using conditional gans,''
\newblock in {\em CVPR}, 2018.

\bibitem{tang2019multichannel}
Hao Tang, Dan Xu, Nicu Sebe, Yanzhi Wang, Jason~J. Corso, and Yan Yan,
\newblock ``Multi-channel attention selection gan with cascaded semantic
  guidancefor cross-view image translation,''
\newblock in {\em CVPR}, 2019.

\bibitem{DBLP:journals/corr/0001T16}
Ming{-}Yu Liu and Oncel Tuzel,
\newblock ``Coupled generative adversarial networks,''
\newblock in {\em NeurIPS}, 2016.

\bibitem{DualGAN}
Zili Yi, Hao Zhang, Ping Tan, and Minglun Gong,
\newblock ``Dualgan: Unsupervised dual learning for image-to-image
  translation,''
\newblock in {\em ICCV}, 2017.

\bibitem{Unet}
Olaf Ronneberger, Philipp Fischer, and Thomas Brox,
\newblock ``U-net: Convolutional networks for biomedical image segmentation,''
\newblock in {\em CVPR}, 2015.

\bibitem{vgg}
Karen Simonyan and Andrew Zisserman,
\newblock ``Very deep convolutional networks for large-scale image
  recognition,''
\newblock in {\em ICLR}, 2015.

\bibitem{salimans2016improved}
Tim Salimans, Ian Goodfellow, Wojciech Zaremba, Vicki Cheung, Alec Radford, and
  Xi~Chen,
\newblock ``Improved techniques for training gans,''
\newblock in {\em NeurIPS}, 2016.

\bibitem{liu2018improved}
Shaohui Liu, Yi~Wei, Jiwen Lu, and Jie Zhou,
\newblock ``An improved evaluation framework for generative adversarial
  networks,''
\newblock {\em arXiv preprint arXiv:1803.07474}, 2018.

\bibitem{che2016mode}
Tong Che, Yanran Li, Athul~Paul Jacob, Yoshua Bengio, and Wenjie Li,
\newblock ``Mode regularized generative adversarial networks,''
\newblock in {\em ICLR}, 2017.

\bibitem{tang2019cycle}
Hao Tang, Dan Xu, Gaowen Liu, Wei Wang, Nicu Sebe, and Yan Yan,
\newblock ``Cycle in cycle generative adversarial networks for keypoint-guided
  image generation,''
\newblock in {\em ACM MM}, 2019.

\bibitem{mechrez2018contextual}
Roey Mechrez, Itamar Talmi, and Lihi Zelnik-Manor,
\newblock ``The contextual loss for image transformation with non-aligned
  data,''
\newblock in {\em ECCV}, 2018.

\bibitem{lin2019refinenet}
Guosheng Lin, Fayao Liu, Anton Milan, Chunhua Shen, and Ian Reid,
\newblock ``Refinenet: Multi-path refinement networks for dense prediction,''
\newblock {\em IEEE TPAMI}, 2019.

\bibitem{Lin:2017:RefineNet}
Guosheng Lin, Anton Milan, Chunhua Shen, and Ian Reid,
\newblock ``Refine{N}et: {M}ulti-path refinement networks for high-resolution
  semantic segmentation,''
\newblock in {\em CVPR}, 2017.

\end{thebibliography}

\end{document}